\documentclass[runningheads]{llncs}

\usepackage{eccv}
\usepackage{eccvabbrv}
\usepackage{amsmath}
\usepackage{amssymb}
\usepackage{booktabs}
\usepackage{graphicx}
\usepackage{placeins}
\usepackage{microtype}
\usepackage{xcolor}
\usepackage{url}
\usepackage{iftex}
\ifPDFTeX
\usepackage[accsupp]{axessibility}
\fi
\usepackage{hyperref}
\usepackage{diagbox}

\newcommand{\method}{Owner3D}

\newcommand{\R}{\mathbb{R}}

\begin{document}

\title{\method{}: Ownership-Guided Style Writing for Training-Free Localized 3D Stylization}
\titlerunning{\method{}}

\author{Suchang Tao \and
Kaifeng Shi \and
Zhiyan Liu \and
Zhuoyuan Jiang \and
Yuqi Ouyang\thanks{Corresponding author.}}

\authorrunning{S.~Tao et al.}

\institute{College of Computer Science, Sichuan University, Chengdu, China\\
\email{\{taosuchang,shikaifeng,liuzhiyan666,jiangzhuoyuan\}@stu.scu.edu.cn yuqi.ouyang@scu.edu.cn}}
\maketitle

\begin{abstract}
Localized 3D stylization aims to modify the appearance of a specified object part while preserving the remaining surfaces. In large reconstruction models (LRMs), this task is challenging because style is injected into intermediate appearance representations before rendering, while compact triplane features are shared across target and non-target surfaces, causing style leakage and boundary ambiguity. We propose \method{}, a training-free framework for localized 3D stylization that integrates localized appearance control directly into the LRM reconstruction process. Specifically, \method{} introduces ownership-guided style writing to restrict reference-style injection to target regions, producing a single localized stylized triplane without additional training while avoiding separate global style and appearance representations. To resolve appearance ambiguity near semantic boundaries, we further introduce boundary dual slots that maintain separate local feature sources for target and non-target regions. Finally, a surface-first texture readout hierarchically combines surface, 3D, and triplane ownership evidence to robustly recover appearance under incomplete visibility. Experiments on a benchmark constructed from Google Scanned Objects and PartNet demonstrate that \method{} consistently outperforms existing 3D stylization methods in target-region style fidelity and non-target appearance preservation, reducing appearance leakage by 86.4\% and 89.9\% compared with StyleSplat and LAENeRF, respectively.

\end{abstract}

\section{Introduction}
\begin{figure*}[!t]
    \centering
    \includegraphics[width=0.95\textwidth]{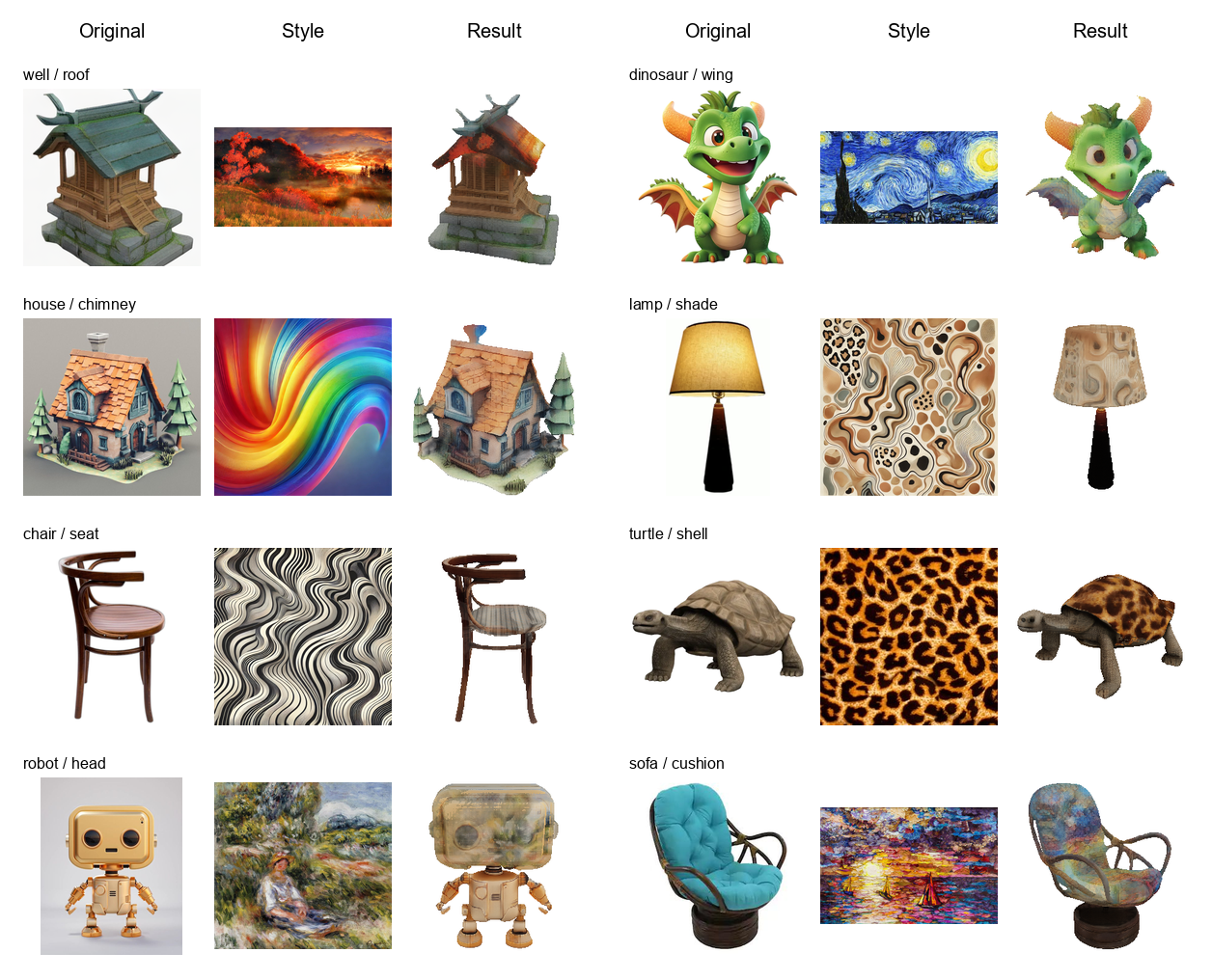}
    \caption{Representative localized 3D stylization results on real-world textured objects from Google Scanned Objects (GSO)~\cite{downs2022googlescannedobjectshighquality}. Each example shows the source object, style reference image, and the stylized 3D output produced by \method{}.}
    \label{fig:eightcase}
\end{figure*}

Feed-forward 3D asset generation has enabled efficient construction of textured 3D assets from images, creating new opportunities for interactive appearance editing. Large reconstruction models (LRMs) further promote this process by mapping one or a sparse set of images to renderable triplane representations in a single forward pass \cite{hong2024lrm,xu2024instantmesh}. Building on this reconstruction capability, recent LRM stylization methods exploit the decoder structure by injecting style-image keys and values into late cross-attention, producing stylized appearance triplanes without training additional 3D models~\cite{oztas2025stylization}. This progress motivates localized 3D stylization, where users can modify the appearance of a specified semantic part, such as a roof, chimney, lampshade, chair seat, or shell, while preserving the remaining object. Representative examples are shown in \cref{fig:eightcase}, where our approach enables localized style transfer to selected regions while preserving the geometry and appearance of non-target regions across diverse objects and styles.

Localized 3D stylization remains challenging due to three issues. First, style information is injected into intermediate appearance representations before rendering, meaning that final-image masking cannot prevent undesired style propagation. Second, compact triplane representations share appearance features across different surface regions, causing target and non-target parts to compete for limited representation capacity near semantic boundaries. Third, incomplete visibility makes it difficult to reliably determine the ownership of each texture query, leading to ambiguous feature selection during texture readout. These challenges require localized stylization to control style writing, resolve boundary ambiguity, and select reliable appearance features.

To address the limitations, \method{} presents a training-free framework for localized 3D stylization that integrates appearance control into the 3D reconstruction process by regulating both ownership-guided style writing and texture readout. Specifically, \method{} introduces ownership-guided style writing in the decoder to localize style injection to the target region, producing a single localized stylized triplane without additional training or separate global style and content representations. Furthermore, \method{} introduces boundary dual slots, a sparse local capacity expansion that separates target and non-target appearance features near shared triplane boundaries without maintaining an additional global content representation. Finally, surface-first texture readout hierarchically integrates surface, 3D, and triplane ownership evidence to select reliable appearance features under incomplete visibility. Our contributions are:

\begin{itemize}
\item We introduce \method{}, a training-free localized 3D stylization framework that enables controllable part-level editing within a feed-forward reconstruction pipeline, achieving high-quality stylization with significantly reduced non-target leakage compared with existing baselines.
\item We establish an ownership-guided style writing paradigm for LRMs that enables localized reference-style injection while maintaining a single localized stylized triplane.
\item We propose boundary dual slots to overcome the appearance ambiguity caused by shared triplane support, enabling independent local feature representation for target and non-target regions.
\item We design surface-first texture readout to robustly recover appearance features by prioritizing surface ownership evidence and progressively leveraging 3D and triplane cues under incomplete visibility.
\end{itemize}

\section{Related Work}

\subsection{3D Appearance Representations}
Neural radiance fields and Gaussian splatting have established effective differentiable 3D appearance representations for high-quality view synthesis \cite{mildenhall2020nerf,kerbl2023gaussian}. To improve the compactness and scalability of 3D representations, triplane architectures encode spatial features on three orthogonal planes, providing an efficient representation scheme for 3D-aware generation and feed-forward reconstruction. Building upon this representation, large reconstruction models (LRMs) reconstruct 3D assets by encoding one or a sparse set of input images into image and triplane tokens, followed by a decoder that produces a reusable appearance triplane in a single forward pass \cite{chan2022eg3d,hong2024lrm,xu2024instantmesh}. However, the compact feature sharing of triplanes introduces ambiguity for localized stylization, as target and non-target regions may rely on overlapping appearance features near semantic boundaries. Different from reconstruction-oriented methods that focus on global appearance recovery, \method{} investigates how compact triplane representations can be controlled for localized part-level stylization.

\subsection{Reference-Guided 3D Stylization}

Classical image stylization transfers reference appearance by optimizing or statistically modulating image features, including neural style transfer, adaptive instance normalization, whitening and coloring transformations, and histogram matching~\cite{gatys2016style,huang2017adain,li2017universal,reinhard2001color}. These effective approaches remain confined to the image domain. 3D stylization extends such transfer to persistent assets represented by meshes, neural radiance fields, Gaussians, or generated multi-view objects~\cite{michel2022text2mesh,huang2022stylizednerf,zhang2022arf,radl2024laenerf,liu2024stylegaussian,song2024style3dattentionguidedmultiviewstyle}. Recent Gaussian methods, including StyleSplat, StylizedGS, and G-Style, improve multi-view stylization but typically require optimized scene representations, mesh-specific processing, or additional feature stages, limiting scalability~\cite{jain2024stylesplat,zhang2024stylizedgs,kovacs2024gstyle}. LRM-based stylization offers a training-free alternative by injecting style keys and values into late decoder cross-attention blocks~\cite{oztas2025stylization}. However, because decoder queries share the same style tokens, this interface is inherently global and struggles with precise part-level stylization. Inspired by attention-based image-editing control~\cite{hertz2023prompt,zhang2023controlnet}, \method{} introduces ownership-guided style writing to localize reference-style injection during LRM reconstruction.

\subsection{Ownership-Guided Local Editing}
Ownership estimation provides essential spatial guidance for localized 3D editing by identifying regions to modify or preserve. Recent promptable segmentation methods obtain object and part masks from image evidence \cite{kirillov2023sam,carion2025sam3}, while related approaches lift 2D ownership information into 3D through multi-view aggregation, open-vocabulary instance proposals, or radiance-field propagation \cite{cen2023segmentanything3d,takmaz2023openmask3d}. However, ownership signals only indicate where an edit should occur and do not determine how appearance changes should be introduced or maintained within a compact 3D representation. Existing instruction-driven 3D editing approaches leverage 2D generative priors to modify 3D assets, but often rely on iterative optimization of scene representations \cite{haque2023instructnerf2nerf}. Recent feed-forward and training-free editors improve efficiency through voxel flows, region-aware merging, or unified scene representations \cite{hu2026easy3efeedforward3dasset,ye2025nano3dtrainingfreeapproachefficient,zhu2026jointedit3dfeedforward3dscene}. Nevertheless, these approaches typically perform editing on reconstructed assets or external representations, rather than controlling how appearance is formed inside a feed-forward reconstruction model. In contrast, \method{} integrates ownership guidance into fixed LRM reconstruction to regulate pre-rendering style writing and appearance readout, mitigating style propagation and unreliable appearance selection under incomplete visibility for efficient part-level stylization.

\section{Method}\label{sec:method}

\begin{figure*}[!t]
    \centering
    \makebox[\textwidth][c]{%
        \includegraphics[width=1\textwidth]{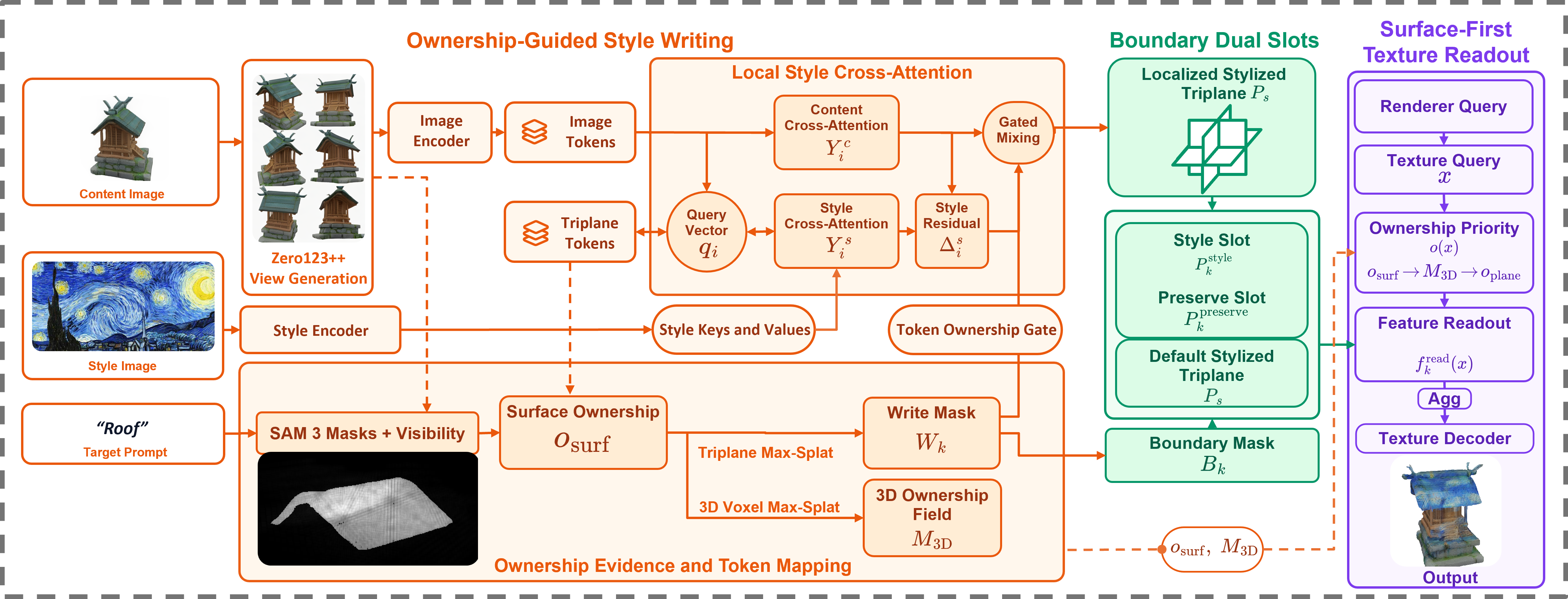}
    }
    \caption{Our framework. Given a content image, a style image, and a target prompt, a frozen LRM first reconstructs the object to estimate surface, 3D, and triplane ownership. Ownership-guided style writing then localizes style injection during appearance decoding to produce a localized stylized triplane. Boundary dual slots provide separate target and preserve features near semantic boundaries, and surface-first texture readout hierarchically integrates ownership evidence to retrieve the appropriate appearance features for rendering.}
    \label{fig:pipeline}
\end{figure*}

\method{} is an LRM-based framework for training-free localized 3D stylization. As shown in \cref{fig:pipeline}, the framework takes a content image, a style image, and a target prompt as inputs. Zero123++~\cite{shi2023zero123plus} expands the content image into six source views, which the frozen LRM encodes into image tokens for triplane reconstruction. The style encoder extracts style features from the style image, while SAM 3~\cite{carion2025sam3} segments the target region in the source views according to the target prompt. A content-only LRM pass first reconstructs the object to recover the geometry, depth, and visibility required for ownership estimation. A second pass then preserves the original content pathway in the early layers and injects ownership-gated style information only into the final 6 appearance-decoder layers, without updating the pretrained model weights. More specifically, the framework consists of three main modules: Ownership-guided style writing restricts style injection to decoder queries associated with the target part, producing a single localized stylized triplane. Boundary dual slots provide separate local style and preserved features where target and non-target surfaces share triplane support. Surface-first texture readout then uses the most reliable available ownership evidence to select the appropriate appearance feature for each surface query before final color decoding. Next, we will elaborate the three core modules.

\subsection{Ownership-Guided Style Writing}\label{subsec:style-writing}

This stage localizes style writing to target-owned decoder queries. A content-only LRM pass first reconstructs the geometry and provides the depth and visibility cues used to estimate surface ownership. The resulting ownership is projected onto the triplanes to gate decoder queries as write mask and lifted into a voxelized field for later texture readout as 3D ownership. A second pass follows the content path in early layers and injects the content-relative style residual only into target-owned queries in the final six appearance layers, producing a localized stylized triplane while preserving the original reconstruction elsewhere.

\noindent\textbf{Surface ownership.}
Let $p\in\R^3$ be a reconstructed surface point and let $\pi_v(p)\in\R^2$ be its projection in view $v$. The view mask $m_v:\R^2\rightarrow[0,1]$ gives a SAM 3 target score. To simplify notations, we use $m_v(p)$ to abbreviate $m_v(\pi_v(p))$. The binary visibility indicator $z_v(p)\in\{0,1\}$ is one only when this projection is visible and passes the depth-consistency test. We estimate whether $p$ belongs to the target part by averaging valid binary mask votes:
\begin{equation}
    o_{\mathrm{surf}}(p)=
    \frac{
        \sum_v z_v(p)\mathbf{1}[m_v(p)\ge\tau_m]
    }{
        \max(1,\sum_v z_v(p))
    }.
    \label{eq:surface-owner}
\end{equation}
Here the 2D mask binarization threshold $\tau_m$ converts each SAM 3 score into a binary target vote. Only depth-consistent visible projections participate in the voting process, and points without any valid visible projection are assigned zero ownership. Thus $o_{\mathrm{surf}}(p)\in[0,1]$ represents the fraction of valid visible views that classify $p$ as belonging to the target part. To construct the 3D ownership field $M_{3D}:\R^3\rightarrow[0,1]$, we max-splat the retained surface ownership scores to their quantized 3D voxel locations. These signals are complementary: surface ownership provides the write masks and preserves visible boundary detail, while $M_{3D}$ supplies spatial support for subsequent texture readout.

\noindent\textbf{Write mask.}
The LRM decoder does not directly write colors to surface points; it writes features to triplane tokens. We therefore map surface ownership to the corresponding triplane cells. Let $\phi_k(p)\in[-1,1]^2$ project $p$ onto plane $k\in\{xy,xz,yz\}$, and let $u_k(p)=\operatorname{round}(\frac{R-1}{2}(\phi_k(p)+1))$ be its quantized coordinate on an $R\times R$ plane. Surface ownership is mapped by a radius-based max-splat:
\begin{equation}
\begin{aligned}
    \overline W_k(u)
    &=
    \max_{\substack{p\in\mathcal S\\
    \lVert u-u_k(p)\rVert_\infty\le r_s}}
    o_{\mathrm{surf}}(p),\\
    g_{\iota(k,u)}
    &=W_k(u),\qquad
    W_k=\mathcal N\!\left(\mathcal S_{\mathrm{2D}}(\overline W_k)\right).
\end{aligned}
\label{eq:write-gate}
\end{equation}
Here the maximum over an empty set is zero, while $\mathcal S_{\mathrm{2D}}$ and $\mathcal N$ denote the spatial smoothing and normalization applied by the implementation. The decoder contains exactly $3R^2$ tokens arranged as three $R\times R$ triplane grids. Therefore, each token has a unique correspondence to a plane index $k$ and a grid coordinate $u$, denoted as $i=\iota(k,u)$. The ownership map $W_k$ thus provides one ownership gate $g_i\in[0,1]$ for each decoder token.

\noindent\textbf{Local style cross-attention.}
Directly applying style cross-attention to all decoder queries would write reference appearance into both target and non-target triplane regions. We therefore use write mask to restrict the style-induced change to target queries. In the late appearance layers, the style encoder provides style keys $k_j^{s,\ell}\in\R^{d_a}$ and values $v_j^{s,\ell}\in\R^C$ for $N_s$ style tokens. For query $i$ with query vector $q_i^\ell\in\R^{d_a}$, the style cross-attention response is:
\begin{equation}
    Y_i^{s,\ell}=\sum_{j=1}^{N_s}
    \operatorname{softmax}\!\left(
    \frac{(q_i^\ell)^{\top}k_j^{s,\ell}}{\sqrt{d_a}}\right)_j
    v_j^{s,\ell}.
    \label{eq:style-attn}
\end{equation}
The same query obtains a content response $Y_i^{c,\ell}\in\R^C$ from the original reconstruction path. We define the content-relative style residual as $\Delta_i^{s,\ell}=Y_i^{s,\ell}-Y_i^{c,\ell}$, which isolates the appearance change introduced by the style reference. Gating this residual preserves the original content response outside the target region while allowing target queries to incorporate style information. Following the layer-wise stylization strategy of 3D Stylization via LRM~\cite{oztas2025stylization}, we apply this operation only to the last few appearance cross-attention layers, which primarily refine appearance. Writing $H_i^\ell\in\R^C$ as the hidden state of query $i$ at layer $\ell$, and $\mathcal{F}_\ell(\cdot)$ as the remaining decoder operations within this layer, \method{} updates each token as:
\begin{equation}
    H_i^{\ell+1}=
    \mathcal F_\ell\!\left(
    H_i^\ell+Y_i^{c,\ell}
    +\alpha g_i\Delta_i^{s,\ell}\right),
    \label{eq:style-writing}
\end{equation}
where $\alpha$ controls the global style strength and $g_i$ denotes the ownership gate derived from the triplane ownership map. When $g_i=0$, the query exactly follows the original content reconstruction path, while increasing $g_i$ progressively incorporates the style residual into target-owned queries. This ownership-gated update confines style writing to the target region and produces a single localized stylized triplane $P_s$ without maintaining separate global style and content triplanes.

\subsection{Boundary Dual Slots}\label{subsec:boundary-slots}

Ownership-guided style writing prevents global style drift, but it does not fully solve a triplane capacity problem. A triplane cell is shared by all 3D points whose projections fall near that cell. Near a semantic boundary, target and preserve surfaces may therefore read from the same local triplane support. To solve the problem, boundary dual slots add a small amount of extra memory only at these ambiguous locations.

Let $u\in\{1,\ldots,H_p\}\times\{1,\ldots,W_p\}$ be a 2D grid coordinate on plane $k\in\{1,2,3\}$ and let $\mathcal{N}_{2D}(u)$ be its radius-$\rho$ neighborhood, where $\rho$ sets the spatial extent used to inspect local ownership. We mark $u$ as a boundary location when this neighborhood contains both target-owned and preserve-owned triplane cells:
\begin{equation}
    B_k(u)=\mathbf{1}\!\left[
    \max_{v\in\mathcal{N}_{2D}(u)}W_k(v)\ge\tau_o
    \;\land\;
    \min_{v\in\mathcal{N}_{2D}(u)}W_k(v)<\tau_o
    \right].
    \label{eq:boundary-slot-mask}
\end{equation}
Here $\tau_o$ separates target-owned from preserve-owned cells, and $B_k\in\{0,1\}^{H_p\times W_p}$ is a sparse boundary mask. It is active only where the local triplane neighborhood crosses the interface between target and preserve regions.

For each active boundary location, we keep two local feature sources. The style slot stores the locally stylized feature from $P_{s,k}$, while the preserve slot stores the corresponding reconstruction feature before style writing. Let $L^c_k\in\R^{H_p\times W_p\times C}$ denote this cached reconstruction feature, with non-boundary entries discarded after slot construction. The product with $B_k$ is broadcast along the channel dimension, and the sparse slots are:
\begin{align}
    P_k^{\mathrm{style}} &= B_k\odot P_{s,k},\\
    P_k^{\mathrm{preserve}} &= B_k\odot L^c_k.
    \label{eq:dual-slots}
\end{align}
Both slots have the same channel dimension $C$ as the triplane, but are used only at active boundary locations. Interior target regions still read from the default stylized triplane, and far-away preserve regions do not allocate extra memory.

\subsection{Surface-First Texture Readout}\label{subsec:surface-readout}

After the stylized triplane and boundary slots are built, the renderer queries appearance features for 3D surface locations. Let $x\in\mathcal{S}\subset\R^3$ be such a texture query. We first decide whether $x$ should behave as target or preserve. Surface ownership from visible source views provides the primary evidence; when it is unavailable, we successively fall back to the 3D ownership field and the triplane write masks:
\begin{equation}
    o(x)=
    \begin{cases}
        o_{\mathrm{surf}}(x), & \text{with valid surface support},\\
        M_{3D}(x), & \text{with valid 3D owner support},\\
        o_{\mathrm{plane}}(x), & \text{otherwise}.
    \end{cases}
    \label{eq:owner-priority}
\end{equation}
Valid surface support requires at least one depth-consistent visible projection, whereas valid 3D support requires $x$ to lie within the observed support of $M_{3D}$. For $x\in\mathcal{S}$, $o_{\mathrm{surf}}(x)$ follows the surface ownership definition above. As a final fallback, we compute $o_{\mathrm{plane}}(x)=\frac{1}{3}\sum_{k=1}^{3}\operatorname{sample}(W_k,\pi_k(x))$ by projecting $x$ onto each plane with $\pi_k$ and bilinearly sampling the corresponding write mask $W_k$.

For plane $k$, let $f_k(x)\in\R^C$ be the feature sampled from the default stylized plane $P_{s,k}$, and let $f_k^{\mathrm{style}}(x),f_k^{\mathrm{preserve}}(x)\in\R^C$ be features sampled from the two boundary slots. We use $b_k(x)=\mathbf{1}[\operatorname{sample}(B_k,\pi_k(x))>0]$ to indicate whether the query falls on a boundary slot. With target threshold $\tau_s$ and preserve threshold $\tau_p$ satisfying $\tau_p<\tau_s$, the per-plane readout is:
\begin{equation}
    f_k^{\mathrm{read}}(x)=
    \begin{cases}
        f_k^{\mathrm{style}}(x), &
        b_k(x)=1\ \land\ o(x)\ge\tau_s,\\
        f_k^{\mathrm{preserve}}(x), &
        b_k(x)=1\ \land\ o(x)\le\tau_p,\\
        f_k(x), & \text{otherwise}.
    \end{cases}
    \label{eq:owner-readout}
\end{equation}
The interval $\tau_p<o(x)<\tau_s$ is intentionally left undecided: at a boundary with uncertain ownership, the default feature $f_k(x)$ avoids a hard switch to either slot. 

The original LRM aggregation operator $\operatorname{Agg}$ then combines the three selected plane features into one query feature $f^{\mathrm{read}}(x)$. Finally, the aggregated feature $f^{\mathrm{read}}(x)$ is passed to the texture decoder to produce the rendered color.

\section{Experiments}\label{sec:experiments}

\subsection{Datasets}

Two datasets are used for the evaluation. The first consists of real-world textured objects from Google Scanned Objects (GSO)~\cite{downs2022googlescannedobjectshighquality}, providing realistic geometry and appearance variations for evaluating localized stylization in scanned assets. The second is derived from PartNet~\cite{mo2018partnetlargescalebenchmarkfinegrained}, whose fine-grained part annotations are adapted to define editable semantic targets. From these 3D assets, we construct a unified evaluation benchmark that supports the input formats required by all the categories of evaluated methods. Source views are rendered using a fixed camera protocol, while a target-part prompt and the corresponding view masks are prepared for each evaluation case. In total, our evaluation includes 49 3D objects and 16 style references.

\subsection{Implementation Details}
We use the publicly released pretrained InstantMesh-Large checkpoint as the frozen reconstruction backbone, following the LRM stylization setting of 3D Stylization~\cite{oztas2025stylization}. \method{} operates entirely at inference time by modifying the late appearance-decoder cross-attention and query-time texture readout, without object-specific fine-tuning. All experiments use single input view, fixed rendering cameras, and the same pretrained InstantMesh parameters. Zero123++~\cite{shi2023zero123plus} synthesizes the source views for reconstruction, while SAM 3 generates the corresponding target-part masks, which are lifted into routing signals for style writing and texture readout. Unless otherwise specified, the 2D mask binarization threshold is $\tau_m=0.5$, the triplane ownership boundary threshold is $\tau_o=0.5$, the boundary neighborhood radius is $\rho=1$, style writing is applied to the last $6$ appearance-decoder cross-attention layers, and the style residual scale is $\alpha=0.90$. For baseline methods, we follow each model's native input requirements and provide the corresponding ground-truth benchmark data rather than restricting all methods to the same single-view input as ours, ensuring a fair comparison across different input modalities. Image-space baselines are evaluated in both single-view and multi-view settings by applying style transfer to individual rendered views or a multi-view image grid. All experiments are conducted on a single NVIDIA RTX 5090 GPU.

% Specifically, we extract features from the relu1\_1, relu2\_1, relu3\_1, and relu4\_1 layers of an ImageNet-pretrained VGG19 network. For a feature map $F_l\in\mathbb{R}^{C_l\times H_lW_l}$, we compute the normalized Gram matrix $G_l=F_lF_l^\top/(C_lH_lW_l)$ and define T-Gram as the mean element-wise squared difference between the target-region and style-reference Gram matrices, averaged over the four layers. The masked target crop is padded by 12 pixels, its non-target pixels are set to white, and both the crop and style reference are resized to $224\times224$ before feature extraction. Because the Gram matrices are normalized and their element-wise errors are averaged, raw T-Gram values are typically on the order of $10^{-4}$. For readability, all tables report T-Gram after multiplication by $10^4$.

\subsection{Evaluation Metrics}
\label{sec:evaluation-metrics}

To evaluate target-region style fidelity, following prior LRM stylization practice~\cite{oztas2025stylization}, we report Style Fidelity and Target Gram Distance (T-Gram). Style Fidelity measures the distance between the channel-wise RGB means and standard deviations of the masked target region and those of the style reference~\cite{reinhard2001color}. T-Gram adopts the Gram-matrix representation commonly used in neural style transfer~\cite{gatys2016style}. To evaluate localization quality and view stability, we report Leakage and Multi-View Consistency (MV Cons.). Leakage measures the appearance discrepancy between paired clean and stylized renders over the non-target foreground~\cite{simsar2025lime}. MV Cons. is computed as the standard deviation of target-region T-Gram across rendered views, averaged over all evaluation cases. We additionally report the average wall-clock runtime for each object-style pair. Lower values indicate better performance for all metrics. More specifically, lower Style Fidelity and T-Gram indicate closer agreement with the reference style, lower Leakage indicates better preservation of non-target appearance, lower MV Cons. indicates more stable stylization across views, and lower runtime indicates higher efficiency.

For the sensitivity and ablation analyses, we additionally report Content SSIM and non-target feature drift. Content SSIM is computed between the stylized and clean renders from the same view and averaged over the target mask; higher values indicate better content preservation. For ablation variant $b$ and evaluation case $n$, let $\widetilde J_n^{(b)}$ and $\widetilde J_n^{(\mathrm{full})}$ denote the corresponding renders restricted to the same non-target foreground mask. We define
\begin{equation}
d_n^{(b)}
=
\frac{1}{2}
\sum_{j\in\{\mathrm{relu3\_1},\mathrm{relu4\_1}\}}
\operatorname{mean}\!\left(
\left|
\psi_j(\widetilde J_n^{(b)})
-
\psi_j(\widetilde J_n^{(\mathrm{full})})
\right|
\right),
\end{equation}
where $\psi_j$ denotes the activation of a fixed ImageNet-pretrained
VGG-19 at layer $j$, and $\operatorname{mean}$ averages over all feature
entries. Across $N$ evaluation cases, we report
$\frac{1}{N}\sum_{n=1}^{N}d_n^{(b)}$ and
$\max_{1\le n\le N}d_n^{(b)}$ as the mean and worst-case non-target
feature drift, respectively. Lower values indicate better non-target
preservation.

\begin{figure*}[!t]
    \centering
    \includegraphics[width=\textwidth]
    {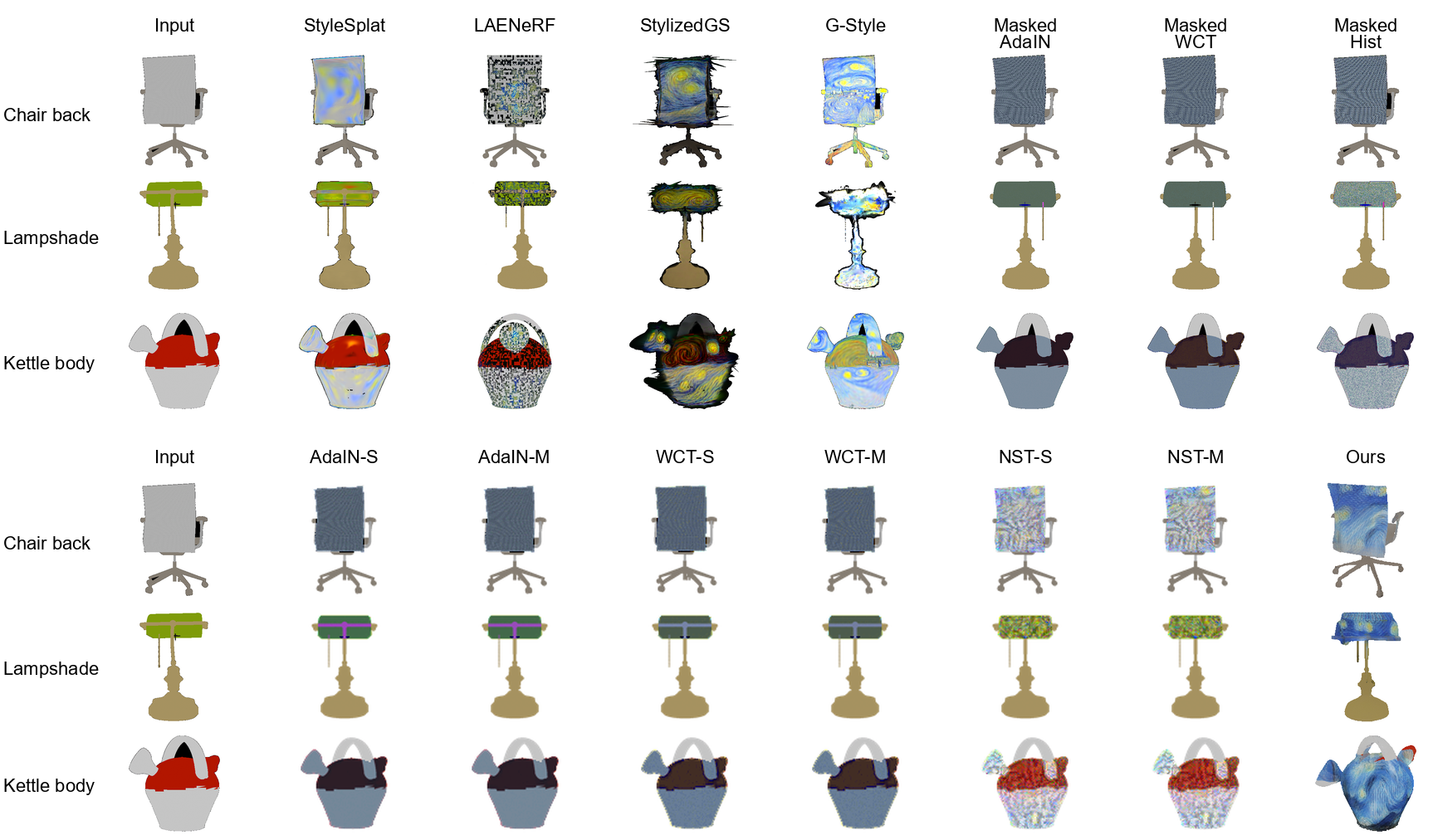}
    \caption{Qualitative comparison on representative objects from the PartNet dataset. The figure compares \method{} with 13 existing 3D and image-space stylization methods.}
    \label{fig:comparison-main}
\end{figure*}

\subsection{Performance Comparisons with Previous Methods}

We compare \method{} with Gaussian-splatting-based 3D stylization, NeRF-based 3D stylization, and masked and unmasked image-space stylization methods on the complete evaluation benchmark. \cref{fig:comparison-main} qualitatively compares localization performance. Existing stylization methods often propagate reference appearance beyond the selected semantic region, particularly around complex boundaries, whereas \method{} confines stylization to the target part while preserving the surrounding appearance.

Tabulated in \cref{tab:comparison-metrics}, \method{} achieves the best Style Fidelity score of $0.1545$ and the lowest comparable Leakage score of $0.0042$, reducing non-target appearance leakage by 86.4\% and 89.9\% compared with StyleSplat and LAENeRF, respectively. Leakage is not reported for masked image-space methods since per-view compositing preserves non-target pixels rather than performing localized 3D editing. Although G-Style achieves a slightly lower T-Gram value of $0.876$ compared with $0.892$ for \method{}, it produces substantially higher Leakage of $0.0608$, indicating that style matching alone does not ensure accurate localized stylization. For view consistency, \method{} achieves an MV Cons. value of $0.00009$, comparable to StyleSplat and lower than other 3D baselines. While some image-space methods report lower MV Cons. or faster runtime, they operate on rendered images and cannot produce reusable view-consistent 3D assets. Among evaluated 3D stylization methods, \method{} is also the most efficient, requiring only $31.10$ seconds per object-style pair. These results demonstrate that \method{} delivers superior localized 3D stylization with high style fidelity, strong appearance preservation, and consistent multi-view rendering.

\renewcommand{\arraystretch}{1.2}
\begin{table*}[!t]
    \centering
    \scriptsize
    \setlength{\tabcolsep}{3.2pt}
    \resizebox{\linewidth}{!}{%
    \begin{tabular}{lccccc}
        \toprule
        Method &
        Style Fidelity $\downarrow$ &
        \shortstack{T-Gram $\downarrow$} &
        MV Cons. $\downarrow$ &
        Leakage $\downarrow$ &
        Runtime $\downarrow$ \\
        \midrule

        \textbf{\method{} (Ours)} &
        \textbf{0.1545} &
        0.892 &
        0.00009 &
        \textbf{0.0042} &
        31.10s \\

        \midrule
        \multicolumn{6}{l}{
        \underline{\textit{Gaussian-Splatting-Based 3D Stylization}}} \\ [3pt]

        StyleSplat~\cite{jain2024stylesplat} &
        0.4909 &
        1.103 &
        0.00010 &
        0.0309 &
        8.15min \\

        StylizedGS~\cite{zhang2024stylizedgs} &
        0.2211 &
        1.058 &
        0.00020 &
        0.0571 &
        1.94min \\

        G-Style~\cite{kovacs2024gstyle} &
        0.6532 &
        \textbf{0.876} &
        0.00029 &
        0.0608 &
        12.52min \\

        \midrule
        \multicolumn{6}{l}{
        \underline{\textit{NeRF-Based 3D Stylization}}} \\ [3pt]

        LAENeRF~\cite{radl2024laenerf} &
        0.4009 &
        0.966 &
        0.00023 &
        0.0417 &
        15.04min \\

        \midrule
        \multicolumn{6}{l}{
        \underline{\textit{Masked Image-Space Stylization}}} \\ [3pt]

        Masked AdaIN~\cite{huang2017adain} &
        0.3032 &
        1.171 &
        0.00023 &
        / &
        19.67s \\

        Masked WCT~\cite{li2017universal} &
        0.3010 &
        1.161 &
        0.00023 &
        / &
        20.00s \\

        Masked HistMatch~\cite{reinhard2001color} &
        0.3767 &
        1.116 &
        0.00021 &
        / &
        28.33s \\

        \midrule
        \multicolumn{6}{l}{
        \underline{\textit{Unmasked Image-Space Stylization}}} \\ [3pt]

        AdaIN (single-view)~\cite{huang2017adain} &
        0.2978 &
        1.219 &
        \textbf{0.00005} &
        / &
        \textbf{0.24s} \\

        AdaIN (multi-view grid) &
        0.2980 &
        1.220 &
        0.00006 &
        / &
        \textbf{0.24s} \\

        WCT (single-view)~\cite{li2017universal} &
        0.2972 &
        1.213 &
        0.00008 &
        / &
        \textbf{0.24s} \\

        WCT (multi-view grid) &
        0.2973 &
        1.213 &
        0.00009 &
        / &
        \textbf{0.24s} \\

        Neural Style Transfer (single-view)~\cite{gatys2016style} &
        0.5041 &
        1.151 &
        \textbf{0.00005} &
        / &
        19.80s \\

        Neural Style Transfer (multi-view grid) &
        0.5037 &
        1.149 &
        \textbf{0.00005} &
        / &
        4.23s \\
        \bottomrule
    \end{tabular}
    }
    \caption{Performance comparison on the PartNet dataset across different 3D and image-space stylization methods. T-Gram values are scaled by $10^4$. Best results are highlighted in bold. Slash marks indicate that leakage is not directly comparable for image-space methods because they preserve non-target pixels through per-view compositing.}
    \label{tab:comparison-metrics}
\end{table*}
\renewcommand{\arraystretch}{1.0}

% The variant
% w/o 3D ownership routing keeps the boundary dual slots but removes the query-level
% owner decision that separates target-side and preserve-side reads. The variant
% w/o 3D write mask disables the target-restricted style-writing mask, allowing
% the style response to be written without the same 3D ownership constraint. The
% variant w/o view-mask support removes the multiview mask evidence used to build
% the routing signal.

\begin{figure}[!t]
    \centering
    \includegraphics[width=\linewidth,height=0.50\textheight,keepaspectratio]{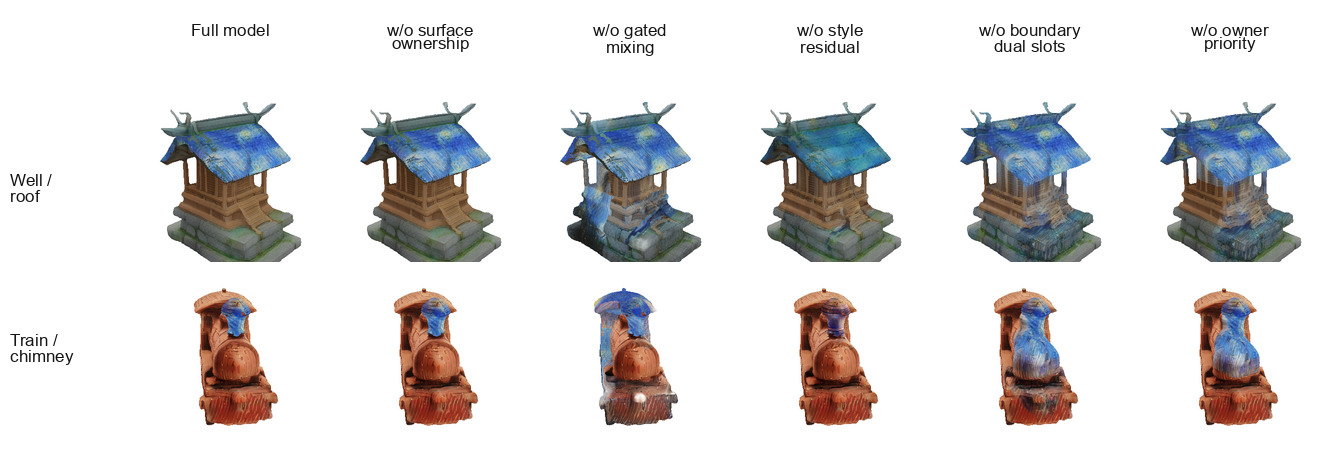}
    \caption{Qualitative ablation study of \method{}. Each variant removes one component to analyze its effect on style localization and boundary appearance preservation.}
    \label{fig:ablation-core}
\end{figure}

\begin{table}[!t]
    \centering
    \small
    \setlength{\tabcolsep}{4pt}
    \renewcommand{\arraystretch}{1.08}
    \begin{tabular}{@{}lccc@{}}
        \toprule
        Variant &
        \shortstack{T-Gram $\downarrow$} &
        \shortstack{Mean Non-Target\\Feature Drift $\downarrow$} &
        \shortstack{Maximum Non-Target\\Feature Drift $\downarrow$} \\
        \midrule
        Full model
            & 0.831 & 0.000 & 0.000 \\
        w/o surface ownership
            & 0.839 & 0.007 & 0.015 \\
        w/o gated mixing
            & 0.843 & 0.374 & 0.558 \\
        w/o style residual
            & 0.950 & 0.133 & 0.293 \\
        w/o boundary dual slots
            & 0.874 & 0.353 & 0.630 \\
        w/o owner priority
            & 0.873 & 0.277 & 0.549 \\
        \bottomrule
    \end{tabular}
    \caption{Performance comparison of ablation variants on the GSO dataset. Each variant removes one component of \method{} to evaluate its contribution. Non-target feature drift measures the VGG-feature deviation of preserved regions from the full model, where lower values indicate better non-target appearance preservation.}
    \label{tab:ablation}
\end{table}

\subsection{Ablation Study}
We perform controlled ablations by individually removing surface ownership, gated mixing, the style residual, boundary dual slots, or owner priority while keeping the input image, style reference, target prompt, camera configuration, pretrained backbone, and renderer fixed. As shown in \cref{fig:ablation-core}, removing gated mixing, boundary dual slots, or owner priority causes the most evident degradation, as these components directly regulate style localization and boundary feature selection. Without gated mixing, style responses leak into preserved regions, while removing boundary dual slots or owner priority introduces ambiguity near shared semantic boundaries. The quantitative results in \cref{tab:ablation} validates these observations. Without gated mixing, T-Gram worsens from $0.831$ to $0.843$, while the mean and maximum non-target feature drift rise to $0.374$ and $0.558$, showing that the ownership gate is primarily responsible for localizing style writing rather than simply increasing style strength. Replacing the content-relative style residual with the direct style response causes the largest degradation in target style matching, increasing T-Gram to $0.950$, which confirms that subtracting the content response avoids an uncontrolled accumulation of appearance features. Boundary dual slots and owner priority are complementary at readout: removing the former yields the largest maximum non-target drift of $0.630$, while removing the latter also causes substantial feature deviation, indicating that boundary-local capacity must be paired with ownership-aware feature selection. By comparison, removing surface ownership produces a smaller change because the remaining 3D and triplane ownership evidence can still provide fallback decisions. Overall, the representative changes show that the proposed components primarily improve localization and non-target appearance preservation while maintaining target-region style transfer.

\begin{figure}[!t]
    \centering
    \includegraphics[width=0.80\linewidth]{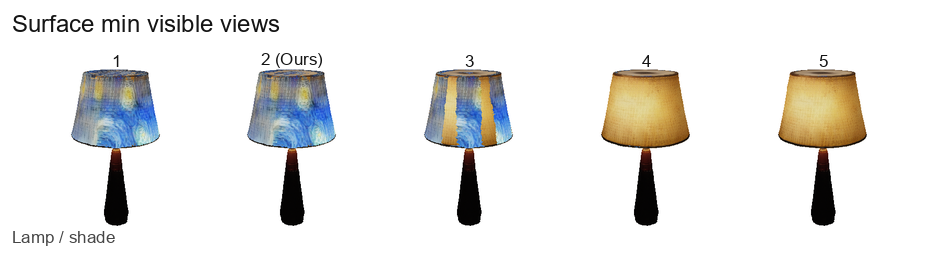}
    \caption{Routing and readout sensitivity on a representative GSO object under different minimum visible-view requirements $K_{\mathrm{vis}}$.}
    \label{fig:sens-visible}
\end{figure}

\subsection{Routing and Readout Sensitivity}
We evaluate the sensitivity of surface-first texture readout to the minimum visible-view requirement $K_{\mathrm{vis}}$, i.e., the minimum number of valid, depth-consistent source-view projections required before accepting surface ownership for a texture query. We vary $K_{\mathrm{vis}}\in{1,2,3,4,5}$ while keeping all other settings fixed, including target-support criteria, style writing, boundary slots, masks, cameras, and renderer. As shown from the lamp-shade case in \cref{fig:sens-visible}, a lower visibility requirement $K_{\mathrm{vis}}$ assigns more queries to the target region, expanding stylization coverage but increasing the risk of ambiguous boundary decisions. In contrast, a higher requirement produces more conservative routing that better protects preserved regions but may reject valid target areas with limited visibility. The default setting of $K_{\mathrm{vis}}=2$ provides a favorable balance by maintaining target coverage while reducing visible leakage, and is therefore selected in this work.

\subsection{Style-Write Sensitivity}

\begin{figure*}[!t]
    \centering
    \includegraphics[width=0.85\textwidth]{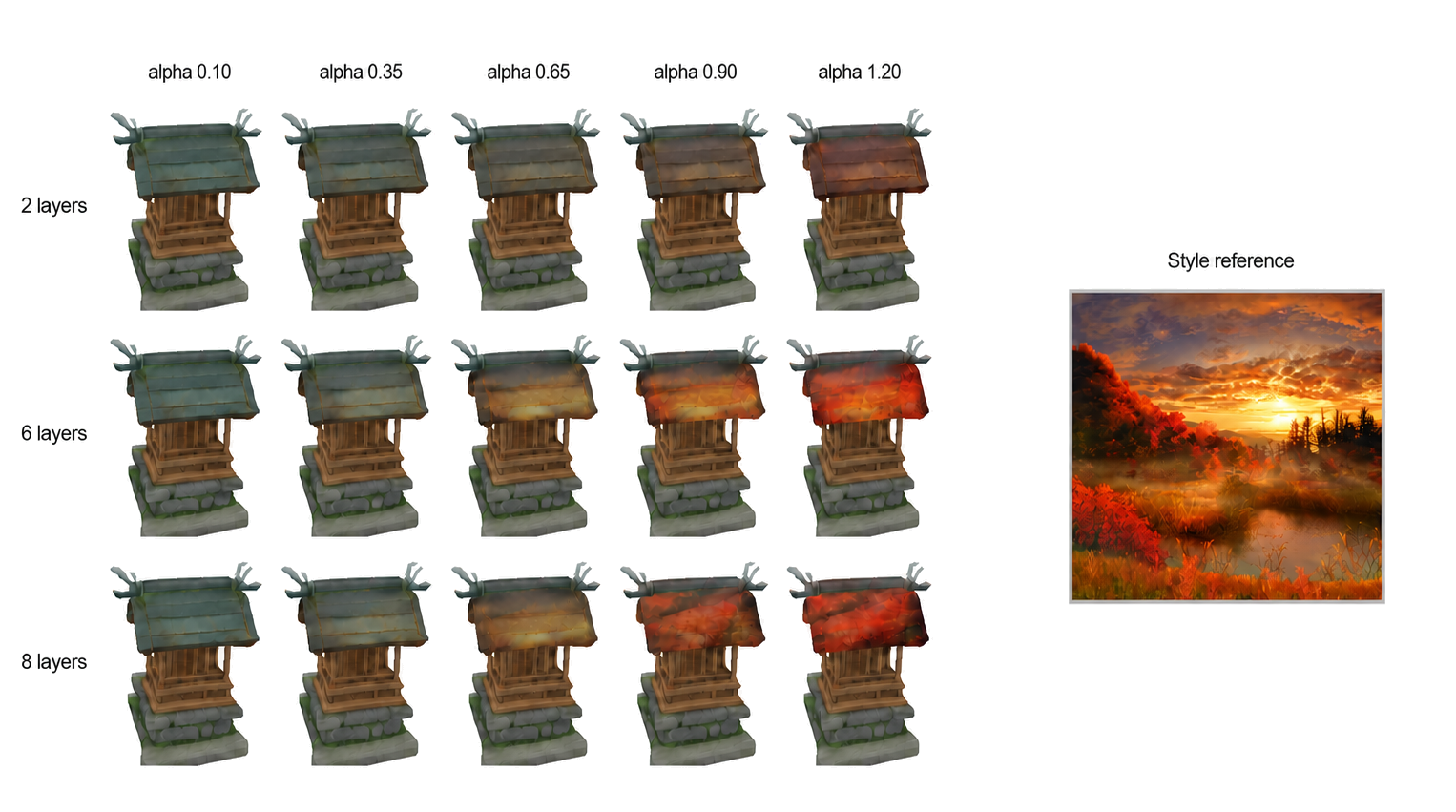}
    \caption{Style-write sensitivity on a representative GSO object. Rows vary the style-write strength $\alpha$, and columns vary the number of style-write layers. Increasing either parameter strengthens style transfer but progressively weakens content preservation.}
    \label{fig:sens-style-write}
\end{figure*}

This experiment isolates style writing inside decoder cross-attention. We vary the number of style-write layers and the write strength while keeping ownership evidence and query-time readout fixed. Each cell in \cref{tab:style-write-sensitivity} reports T-Gram, Style Fidelity, and Content SSIM on the stylized target region from the same view demonstrated in \cref{fig:sens-style-write}. The default setting of $\alpha=0.90$ and six-layer is a relative optimum on the observed style--content trade-off rather than the optimum of every individual metric. Increasing $\alpha$ to $1.20$ further improves T-Gram from $0.60$ to $0.52$ and Style Fidelity from $0.079$ to $0.066$, but reduces Content SSIM from $0.751$ to $0.714$. The qualitative comparison in \cref{fig:sens-style-write} provides an additional spatially localized criterion that is not captured by these global style statistics. As the style reference retains a coherent spatial layout rather than being spatially shuffled, the selected roof region can be aligned with the sunset-colored patch at the corresponding relative position. Under this strict alignment, $\alpha=0.90$ with 6 layers reproduces the locally corresponding color more faithfully, whereas $\alpha=1.20$ or 8 layers applies a stronger global style response at the expense of this local correspondence and source structure. Hence, $\alpha=0.90$ and 6 layers are selected in this work.

\begin{table}[!t]
    \centering
    \scriptsize
    \setlength{\tabcolsep}{4.0pt}
    \renewcommand{\arraystretch}{1.10}
    \resizebox{\linewidth}{!}{%
    \begin{tabular}{cccc}
        \toprule
        \multicolumn{4}{c}{
            Cell: T-Gram ($\times 10^4$) $\downarrow$ /
            Style Fidelity $\downarrow$ /
            Content SSIM $\uparrow$
        } \\
        \midrule
        \diagbox[width=7.2em,height=2.8em]{$\alpha$}{Layers}
        & 2 & 6 & 8 \\
        \midrule
        0.10
        & 0.81 / 0.212 / 0.998
        & 0.81 / 0.211 / 0.993
        & 0.81 / 0.211 / 0.988 \\
        0.35
        & 0.79 / 0.199 / 0.986
        & 0.79 / 0.189 / 0.940
        & 0.79 / 0.190 / 0.921 \\
        0.65
        & 0.75 / 0.181 / 0.955
        & 0.72 / 0.132 / 0.813
        & 0.74 / 0.135 / 0.804 \\
        0.90
        & 0.71 / 0.166 / 0.917
        & 0.60 / 0.079 / 0.751
        & 0.63 / 0.113 / 0.740 \\
        1.20
        & 0.65 / 0.144 / 0.858
        & 0.52 / 0.066 / 0.714
        & 0.55 / 0.100 / 0.706 \\
        \bottomrule
    \end{tabular}%
    }
    \caption{
        Two-dimensional style-write sensitivity. Bold indicates the
        best value for each metric; red marks the selected operating point.
    }
    \label{tab:style-write-sensitivity}
\end{table}

\section{Conclusion}
We presented \method{}, a training-free localized 3D stylization framework that integrates appearance control into the LRM reconstruction process. Ownership-guided style writing localizes reference-style injection at decoder queries and produces a single stylized triplane while preserving non-target appearance. Boundary dual slots and surface-first texture readout further address feature ambiguity near semantic boundaries and improve appearance selection under incomplete visibility. Extensive experiments demonstrate our reduced appearance leakage and high-fidelity target stylization compared with existing baselines, providing a practical solution for controllable appearance editing in large reconstruction models. Our future work will explore improved region control and richer appearance representations for 3D stylization.

\bibliographystyle{splncs04}
\bibliography{main}

\end{document}